\documentclass{article}

\usepackage[main, final]{neurips_2026}

\usepackage[utf8]{inputenc} 
\usepackage[T1]{fontenc}    
\usepackage{hyperref}       
\usepackage{url}            
\usepackage{booktabs}       
\usepackage{amsfonts}       
\usepackage{nicefrac}       
\usepackage{microtype}      
\usepackage{xcolor}         
\usepackage{amsmath}
\usepackage{enumitem}
\usepackage{graphicx} 
\usepackage[table]{xcolor}
\usepackage[most]{tcolorbox}
\usepackage{algorithm}
\usepackage{algpseudocode}
\usepackage{subcaption}

\title{Drift-Aware Multimodal User Representation Learning via Multi-Scale Temporal Modeling and Sparse Mixture-of-Experts}

\author{%
  \textbf{Ziqing Qian}$^{1,2}$ \qquad
  \textbf{Haohang Chen}$^{1}$ \qquad
  \textbf{Shengqi Dang}$^{1,2,3}$ \qquad
  \textbf{Yuhan Xiong}$^{1}$ \\
  \textbf{Canyu Shen}$^{1}$ \qquad
  \textbf{Jiaying Lei}$^{1,2,3}$ \qquad
  \textbf{Nan Cao}$^{1,2,3}$\thanks{Corresponding author.} \\[0.5em]
  \texttt{qianziqing0719@gmail.com, nan.cao@tongji.edu.cn} \\
  $^{1}$Intelligent Design \& InnoVation Lab, Tongji University, Shanghai, China \\
  $^{2}$Shanghai Research Institute for Intelligent Autonomous System, Tongji University, Shanghai, China\\
  $^{3}$Shanghai Innovation Institute, Shanghai, China
}

\def\oursName{DUMoE}
\begin{document}

\maketitle

\begin{abstract}
Understanding user preferences from noisy and temporally evolving social media behaviors is fundamentally challenging due to interest drift, where user preferences shift across time and exhibit both multi-scale temporal patterns and diverse co-existing interests. To address this, we propose \textbf{\oursName}, a unified framework for drift-aware multimodal user representation learning. Our model consists of (i) a temporal dynamics-aware backbone that captures and integrates static profiles, short-term behavioral signals, and long-term dependencies into a coherent representation, and (ii) a sparse mixture-of-experts (MoE) interest adapter that disentangles multiple latent interests via expert specialization and adaptive routing. Each expert models a distinct interest subspace, while a gating network dynamically selects and aggregates a sparse subset of relevant experts for each user. To enable stable and effective optimization, we further introduce a three-stage training strategy that decouples backbone learning, expert specialization, and gating optimization. Extensive experiments on real-world social media datasets show that \textbf{\oursName} consistently outperforms state-of-the-art methods on both user interest prediction and interaction prediction tasks.
Code is available at \url{https://github.com/idvxlab/DUMoE}.
\end{abstract}
\section{Introduction}
Modeling user preferences is a fundamental problem in machine learning, underpinning applications such as recommendation \cite{kang2018self}, personalization \cite{hidri2024learning}, and information diffusion \cite{kempe2003maximizing,qian2026designed}.
A core challenge arises from interest drift, where user preferences evolve over time, exhibiting both long-term stability and short-term dynamics, while simultaneously spanning multiple co-existing interests. Effectively capturing such multi-scale temporal dynamics and multi-interest structure is essential for learning adaptive and robust user representations.

Social media platforms (e.g., X) provide a continuous stream of multimodal user-generated data, including textual content, visual signals, and interaction behaviors, offering a valuable testbed for modeling fine-grained user preference evolution. However, effectively leveraging such data remains challenging due to (1) heterogeneous multimodal inputs, (2) the need to capture multi-scale temporal dynamics, and (3) data sparsity with irregular user activity. Existing user profiling methods typically rely on static representations~\cite{ sun2019bert4rec, yuan2020parameter} or coarse temporal modeling~\cite{hidasi2015session, tang2018personalized, wu2020ptum}, limiting their ability to capture interest drift. Moreover, most approaches focus on single modalities, overlooking the complementary signals across modalities. Critically, prior work lacks a unified framework that jointly models multi-modal information, multi-scale temporal dynamics, and both explicit (interpretable) and implicit (latent) user preferences.

In this work, we propose \oursName, a \textbf{\underline{D}}rift-aware \textbf{\underline{U}}ser \textbf{\underline{M}}odeling framework with \textbf{\underline{M}}ixture-\textbf{\underline{O}}f-\textbf{\underline{E}}xperts. This technique learns hierarchical user representations by jointly modeling multimodal signals, multi-scale temporal dynamics, and user interest structure within a unified architecture.
We first encode heterogeneous user-generated content into a shared semantic space, enabling consistent cross-modal representations. On top of this, we decompose user preferences into two complementary components: (1) explicit preference distributions over predefined interest domains, and (2) latent representations that capture both short-term behavioral dynamics and long-term dependencies. Temporal evolution is modeled via a multi-scale design, combining sequential modeling of recent activities with global aggregation over historical interactions.
To further capture diverse user intents, we introduce a multi-interest module based on sparse mixture-of-experts, which disentangles user interests into specialized subspaces through expert specialization and adaptive routing. Extensive experiments on real-world datasets demonstrate that our approach consistently outperforms strong baselines in modeling user preference dynamics and adapting to evolving interests. Our contributions are summarized as follows:
\begin{itemize}[leftmargin=1.0em]
\item \textbf{Unified drift-aware user representation learning framework.} 
We propose a unified framework that jointly models \textbf{multimodal signals}, \textbf{multi-scale temporal dynamics}, and both \textbf{explicit and latent user preferences}. The proposed backbone integrates static profiles, short-term behaviors, and long-term dependencies into a hierarchical representation, enabling effective modeling of evolving user interests.

\item \textbf{Sparse mixture-of-experts for multi-interest disentanglement.}
We introduce a \textbf{sparse mixture-of-experts (MoE) interest adapter} that decomposes user representations into specialized interest subspaces via expert specialization and adaptive routing. The design promotes \textbf{disentangled, interpretable, and personalized} multi-interest modeling.

\item \textbf{Large-scale temporally annotated multimodal dataset.} 
We construct a large-scale dataset from X with \textbf{chronologically evolving interest annotations}, covering 15 domains with 14K+ users, 7.7M tweets, and 2.9M images, providing a realistic benchmark for modeling user interest drift.
\end{itemize}

\section{Related Work}

\paragraph{Multimodal Representation Learning.} Modeling user behavior over social media requires representations that effectively integrate heterogeneous signals, particularly textual and visual content~\citep{purificato2024user,eke2019survey}. Cross-modal alignment architectures such as VisualBERT~\citep{li2019visualbert} and ViLBERT~\citep{lu2019vilbert} enable fine-grained modality interactions via region-level grounding. Contrastive frameworks including CLIP~\citep{radford2021learning} and ALBEF~\citep{li2021align} scale alignment to web-level data, and generative models such as Flamingo~\citep{alayrac2022flamingo}, BLIP-2~\citep{li2023blip}, and LLaVA~\citep{liu2023visual} extend multimodal learning to complex reasoning. However, these models are designed for curated, domain-specific corpora rather than the noisy, temporally irregular, and user-attributed streams characteristic of social media---they produce instance-level representations with no mechanism to aggregate signals across a user's evolving behavioral history.

\paragraph{Sequential User Behavior Modeling.}
Sequential modeling aims to capture temporal dynamics in user behavior, but existing approaches largely operate at the item level with limited semantic expressiveness. Early methods such as GRU4Rec~\citep{hidasi2015session} and Caser~\citep{tang2018personalized} model short-term dependencies using recurrent and convolutional architectures. Transformer-based models, including SASRec~\citep{kang2018self} and BERT4Rec~\citep{sun2019bert4rec}, improve long-range dependency modeling via self-attention mechanisms. PTUM~\citep{wu2020ptum} further pre-trains user representations from unlabeled behavior sequences. Structural and contrastive enhancements are introduced in SR-GNN~\citep{wu2019session} and DuoRec~\citep{qiu2022contrastive}, while UniSRec~\citep{hou2022towards}, VQ-Rec~\citep{hou2023learning}, and HM4SR~\citep{zhang2025hierarchical} incorporate textual or multimodal item features. Yet these methods model behavior at the item level and largely neglect user-level multi-scale temporal dynamics, and open-domain interest drift.

\paragraph{User Interest Disentanglement.}
Single-vector representations fail to capture preference diversity~\citep{hidri2024learning}, motivating multi-interest modeling. MIND~\citep{li2019multi} and ComiRec~\citep{cen2020controllable} extract multiple interest vectors via capsule routing and self-attention, while S3-Rec~\citep{zhou2020s3}, DCCF~\citep{ren2023disentangled}, and SINE~\citep{tan2021sparse} improve disentanglement through mutual information, graph contrastive learning, and sparse prototypes, respectively. DisMIR~\citep{du2024disentangled} addresses interest collapse via spectral clustering, and HORAE~\citep{hu2025horae} incorporates relative temporal intervals to track interest shifts over time. Despite this, interest subspaces are learned at a fixed snapshot: how their composition and relative salience drift as user behavior evolves over time remains unmodeled.

\paragraph{Summary.}
In summary, prior work has made substantial progress along three complementary directions: multimodal representation learning, sequential modeling, and interest disentanglement. However, these lines of research are largely developed in isolation. Existing multimodal methods overlook temporal dynamics, sequential models underutilize multimodal signals, and disentanglement approaches rarely consider both temporal evolution and heterogeneous content. To bridge these gaps, we propose a unified framework that jointly models multimodal signals, temporal dynamics, and multi-faceted user interests, enabling dynamic and expressive user representations for evolving social media environments.




\section{Problem Formulation}

Given a set of users $\mathcal{U}$, each user $u \in \mathcal{U}$ is associated with a profile $\mathbf{p}_u$ and a chronologically ordered interaction sequence $\mathcal{S}_u$. Our goal is to learn a mapping function
\begin{equation}
f: (\mathcal{S}_u, \mathbf{p}_u) \mapsto \mathbf{h}_u \in \mathbb{R}^d,
\end{equation}
where $\mathbf{h}_u$ denotes the learned user representation that captures three key aspects of a user's behavior: 
(1) \textbf{interaction-level preference signals}, 
(2) \textbf{multi-scale temporal dependencies} that reflect evolving behavioral dynamics, and 
(3) \textbf{multi-interest structure} across diverse user activities. Here, the interaction sequence $\mathcal{S}_u$ is formally defined as
\begin{equation}
\mathcal{S}_u = \{{(m_1^{u}, t_1^{u}), (m_2^{u}, t_2^{u}), \dots, (m_{n}^{u}, t_{n}^{u})}\},
\end{equation}
where \(t_i^{u}\) denotes the timestamp of the i-th interaction, with \(t_1^{u}<t_2^{u}<\dots<t_{n}^{u}\). Each \(m_i^{u}\in\mathcal{M}\) represents the multimodal content associated with the interaction (e.g., posted or retweeted text/image content). The user profile $\mathbf{p}_u$ consists of static attributes, including numerical features (e.g., follower counts) and textual descriptions (e.g., self-declared biography).

\section{Methodology}
\label{app:method}

\begin{figure}[t]
    \centering
    \includegraphics[width=1.0\linewidth]{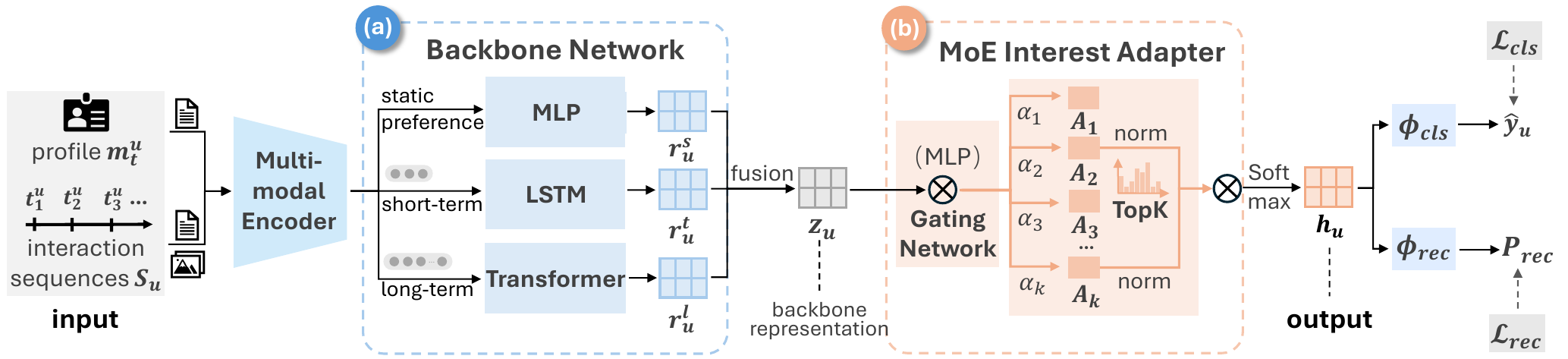}
    \caption{Algorithm Overview.}
    \label{fig:pipeline}
    \vspace{-1.0em}
\end{figure}

In this section, we first present the proposed representation learning framework and then introduce the loss function and the stage-wise optimization strategy. The framework (Fig.~\ref{fig:pipeline}) consists of two key components: a \textbf{temporal dynamics-aware backbone network} and a \textbf{sparse mixture-of-experts (MoE) interest adapter}. 
The backbone network (Fig.~\ref{fig:pipeline}(a)) learns hierarchical user representations by capturing both short-term behavioral dynamics and long-term interaction dependencies. Built upon this representation, the MoE interest adapter (Fig.~\ref{fig:pipeline}(b)) performs \textbf{multi-interest disentanglement} through expert specialization and adaptive routing, enabling dynamic preference modeling. 

\subsection{Temporal Dynamics-Aware Backbone Network}
\label{sec:backbone}
As shown in Fig.~\ref{fig:pipeline}(a), the backbone network is designed to capture user preferences by jointly leveraging static profile information and the user's temporally evolving interaction behaviors. The architecture consists of two main components:

\paragraph{Post and Profile Encoding.}
We first encode both user posts and profile information into a unified embedding space based on a multimodal encoder. Here, we adopt a pre-trained UNITE encoder~\cite{kong2025modality} due to its strong capability in learning aligned text-image representations. For each user post $m_t^{u}$ consisting of textual and visual content, we apply the encoder $\mathcal{E}$ to jointly encode both modalities into a post representation $\mathbf{e}_t^{u}$:
\begin{equation}
    \mathbf{e}_t^{u}=\mathcal{E}(m_t^{u})
\end{equation}
In addition, we also encode the heterogeneous attributes of a user profile $\mathbf{p}_u$ separately: numerical features (e.g., follower count, age) are normalized via min--max scaling and projected into embeddings, while textual attributes (e.g., self-declared biography) are encoded using $\mathcal{E}$. The final profile representation $\mathbf{a}_u$ is constructed by concatenating all attribute embeddings.


\paragraph{Temporal Dynamic Preference Modeling.}

Next, to capture the diverse behavioral patterns exhibited by users in real-world social media environments, we model user preferences from three complementary temporal perspectives: (1) static preference signals, (2) short-term dynamic patterns, and (3) long-term interests. Accordingly, we extract three corresponding feature representations, denoted by \(\mathbf{r}_u^{s}\), \(\mathbf{r}_u^{t}\), and \(\mathbf{r}_u^{l}\), and integrate them into a unified user representation.
These features are defined as
\begin{align}
    \mathbf{r}_u^{s} &= \phi_{s}(\mathbf{a}_u), \label{eq:sta}\\
    \mathbf{r}_u^{t} &= \phi_{t}(\mathbf{S}_u^{t}), \label{eq:short}\\
    \mathbf{r}_u^{l} &= \phi_{l}(\mathbf{S}_u^{l}), \label{eq:long}
\end{align}
where \(\phi_{s}(\cdot)\) is a two-layer MLP that encodes the static profile feature \(\mathbf{a}_u\); \(\phi_{t}(\cdot)\) is an LSTM that processes the short-term interaction sequence \(\mathbf{S}_u^{t}\) to capture recent behavioral dynamics; and \(\phi_{l}(\cdot)\) is a Transformer encoder that models long-range dependency patterns over the long-term interaction sequence \(\mathbf{S}_u^{l}\).
The short-term and long-term sequences are defined as
\begin{equation}
    \mathbf{S}_u^{t} = \{\mathbf{e}_{i}^{u} \mid i = T_u - L_t + 1, \dots, T_u\},
\end{equation}
\begin{equation}
    \mathbf{S}_u^{l} = \{\mathbf{e}_{i}^{u} \mid i = T_u - L_l + 1, \dots, T_u\},
\end{equation}
where \(L_{t}\) and \(L_{l}\) denote the lengths of the short-term and long-term windows, respectively, with \(L_{t} << L_{l}\), and \(T_u\) is the current time step for user \(u\).

Finally, we fuse the three features through direct sum followed by nonlinear projection:
\begin{equation}
\label{eq:fusion}
    \mathbf{z}_u = \phi_{f}\left(\mathbf{r}_u^{s} \oplus \mathbf{r}_u^{t} \oplus \mathbf{r}_u^{l}\right),
\end{equation}
where \(\oplus\) denotes the direct sum (concatenation) operation, and \(\phi_{f}(\cdot)\) is an MLP that integrates the three components into a unified backbone representation \(\mathbf{z}_u\).

\subsection{Mixture-of-Experts Interest Adapter}
To capture multiple latent interests underlying diverse user behaviors, we introduce a \textbf{sparse Mixture-of-Experts (MoE) interest adapter}. The module consists of a set of lightweight expert networks and an adaptive gating mechanism. Each expert is implemented as an adapter module that specializes in modeling a distinct latent interest subspace, enabling \textbf{multi-interest disentanglement}. The gating network learns a user-specific routing distribution over experts, dynamically selecting a sparse subset of relevant experts for each user.  
The final user representation is obtained via a gated aggregation of expert outputs:
\begin{equation}
    \mathbf{h}_{u} = \sum_{k=1}^{K} \alpha_{u,k}\,\mathcal{A}_k(\mathbf{z}_u),
    \label{eq:moegate}
\end{equation}
where $K$ denotes the number of experts, $\mathcal{A}_k(\cdot)$ is the $k$-th interest expert, and $\alpha_{u,k}$ is the gating weight for user $u$. The gating weights satisfy $\sum_{k=1}^{K} \alpha_{u,k} = 1$. In practice, we employ sparse routing, where only a subset of experts receives non-zero weights. Next, we detail the expert design and gating mechanism below.

\paragraph{Interest Expert.}
Each expert is implemented as a lightweight adapter that transforms the shared backbone representation $\mathbf{z}_u$. Specifically, the $k$-th expert is defined as
\begin{equation}
    \mathcal{A}_k(\mathbf{z}_u) = \mathbf{z}_u + \phi_k(\mathbf{z}_u),
\end{equation}
where $\phi_k(\cdot)$ is a multi-layer perceptron (MLP) that models interest-specific transformations. The residual formulation allows each expert to capture an \textbf{interest-specific deviation} from the shared representation, ensuring that all experts remain anchored in a common semantic space while enabling specialization across distinct interest dimensions.

\paragraph{Gating Weight.}
To mitigate interference across interest subspaces and encourage expert specialization, we adopt a \textbf{sparse top-$K'$ routing} mechanism. The gating weights are defined as
\begin{equation}
    \alpha_{u,k} =
    \begin{cases}
    \dfrac{\tilde{\alpha}_{u,k}}{\sum_{k' \in \mathcal{K}_u} \tilde{\alpha}_{u,k'}} & k \in \mathcal{K}_u \\[6pt]
        0 & k \notin \mathcal{K}_u
    \end{cases}
\end{equation}
where $\mathcal{K}_u$ contains the indices of the top-$K'$ experts selected from the routing scores $\tilde{\boldsymbol{\alpha}}_u$. 
This design enforces sparse activation, allowing each user to be represented by a small subset of dominant interests, thereby improving both efficiency and interpretability.
The routing scores are computed as
\begin{equation}
\tilde{\boldsymbol{\alpha}}_u = \mathrm{softmax}(\mathbf{W}_g \mathbf{z}_u + \mathbf{b}_g),
\end{equation}
where $\mathbf{W}_g$ and $\mathbf{b}_g$ are learnable parameters.

\subsection{Learning Objective and Training Procedure}
\label{app:method 4.3}
To enable the learned representation to understand user preferences and behaviors, we employ a supervised learning approach to learn the representation. We use two learning objectives: (1) interest classification and (2) interaction prediction. The overall training objective is a sum of the two loss terms for each task, corresponding to the two tasks respectively:
\begin{equation}
    \mathcal{L} = \mathcal{L}_{\mathrm{cls}} + 
    \lambda\,\mathcal{L}_{\mathrm{rec}}
\end{equation}
where $\mathcal{L}_{\mathrm{cls}}$ is the interest classification loss, $\mathcal{L}_{\mathrm{rec}}$ is the interaction prediction loss, and $\lambda$ is a weight balances the two parts.

\paragraph{Interest Classification.}
The learning objective for interest classification minimizes a 
class-weighted focal loss between the ground-truth interest label 
$y_u$ and the prediction $\phi_{\mathrm{cls}}(\mathbf{h}_u)$:
\begin{equation}
    \mathcal{L}_{\mathrm{cls}} = -\frac{1}{B_{\mathrm{cls}}}\sum_{i=1}^{B_{\mathrm{cls}}}
    w_{y_i}(1-p_{y_i})^{\gamma}\log p_{y_i}
\end{equation}
Here, $p_{y_i} = \phi_{\mathrm{cls}}(\mathbf{h}_u)_{y_i}$ denotes the predicted probability of the ground-truth class, where $i$ indexes users in a mini-batch. $B_{\mathrm{cls}}$ is the batch size of users for the classification task. $w_{y_i}$ is the inverse-frequency class weight for handling class imbalance, and $\gamma=2$ is the focusing parameter. $\phi_{\mathrm{cls}}(\cdot)$ is implemented as $\mathrm{softmax}(\mathbf{W}_{\mathrm{out}}\mathbf{h}_u+\mathbf{b}_{\mathrm{out}})$ with learnable parameters $\mathbf{W}_{\mathrm{out}},\mathbf{b}_{\mathrm{out}}$, and $y_u$ is assigned from the $C$ predefined interest categories based on the user's interaction history.

\paragraph{Interaction Prediction.}
For interaction prediction, we aim to learn user representations such that the representation of a user is similar to the representation of the next post they actually interact with (e.g., a post or repost), and dissimilar to randomly sampled negative posts. To achieve this, we adopt a negative sampling approach with binary cross-entropy (BCE) loss, which encourages high inner products for positive pairs and low inner products for negative pairs, defined as:
\begin{equation}
    \mathcal{L}_{\mathrm{rec}} = -\frac{1}{B_{\text{rec}}} \sum_{i=1}^{B_{\text{rec}}} \left[ \log \sigma(s_i^+) + \frac{1}{K} \sum_{k=1}^{K} \log \sigma(-s_{ik}^-) \right]
\end{equation}
where \(\sigma(\cdot)\) is the sigmoid function, \(B_{\text{rec}}\) is the mini-batch size consisting of users who have interaction samples, and \(K\) denotes the number of negative samples per user. For each user \(i\) in this batch, \(\mathbf{h}_u\) denotes the representation of user \(i\), and \(s_i^+ = \phi_{\mathrm{rec}}(\mathbf{h}_u)^\top \mathbf{v}_i^+\) is the inner product between the user's projected representation \(\phi_{\mathrm{rec}}(\mathbf{h}_u)\) and the positive post embedding \(\mathbf{v}_i^+\) (the ground-truth next post). Here, \(\phi_{\mathrm{rec}}\) is implemented as an MLP. Similarly, \(s_{ik}^- = \phi_{\mathrm{rec}}(\mathbf{h}_u)^\top \mathbf{v}_{ik}^-\) is the inner product with the \(k\)-th negative post embedding, where \(\{\mathbf{v}_{ik}^-\}_{k=1}^{K}\) are sampled negative posts for user \(i\).

\paragraph{Training Strategy.}
We adopt a three-stage optimization procedure to stabilize training and promote expert specialization: (1) \textbf{backbone pre-training}, (2) \textbf{expert specialization}, and (3) \textbf{gating optimization}. The prediction heads $\phi_{\mathrm{cls}}$ and $\phi_{\mathrm{rec}}$ are jointly optimized across all stages.
\textbf{Stage 1: Backbone Pre-training.}
We train the backbone network together with the prediction heads, while removing the MoE adapter. The user representation is directly taken as $\mathbf{h}_u = \mathbf{z}_u$. This stage learns a strong and discriminative shared representation.
\textbf{Stage 2: Expert Specialization.}
We freeze the backbone and train each expert independently using users whose dominant interest label matches the expert index. This encourages each expert to capture a distinct interest subspace. After this stage, the expert pool $\{\mathcal{A}_1, \dots, \mathcal{A}_K\}$ consists of specialized and fixed experts.
\textbf{Stage 3: Gating Optimization.}
We freeze both the backbone and the experts, and train only the gating network parameters ($\mathbf{W}_g$, $\mathbf{b}_g$), allowing it to learn a stable user-specific routing distribution over the specialized experts.

\textbf{Inference.}
At inference time, the full model is applied end-to-end. The backbone and adapter representations are further combined via a KL-based gating signal (see Appendix ~\ref{app:kl_gating} for details).

This stage-wise design decouples representation learning, expert specialization, and routing optimization, resulting in improved stability, clearer expert roles, and more reliable personalized routing.



\section{Experiments}
\label{app:experiments}
To evaluate the effectiveness of the proposed technique, we first introduce a large-scale multimodal dataset collected from X (detailed in Section~\ref{sec:dataset}). Based on this dataset, we detail the experimental settings and conduct comprehensive experiments on two downstream tasks: (1) user interest classification and (2) social interaction prediction, to assess the quality and utility of the learned representations. Finally, we present ablation studies to analyze the contribution of each key component.

\begin{figure}[htbp]
    \centering
\includegraphics[width=1.0\linewidth]{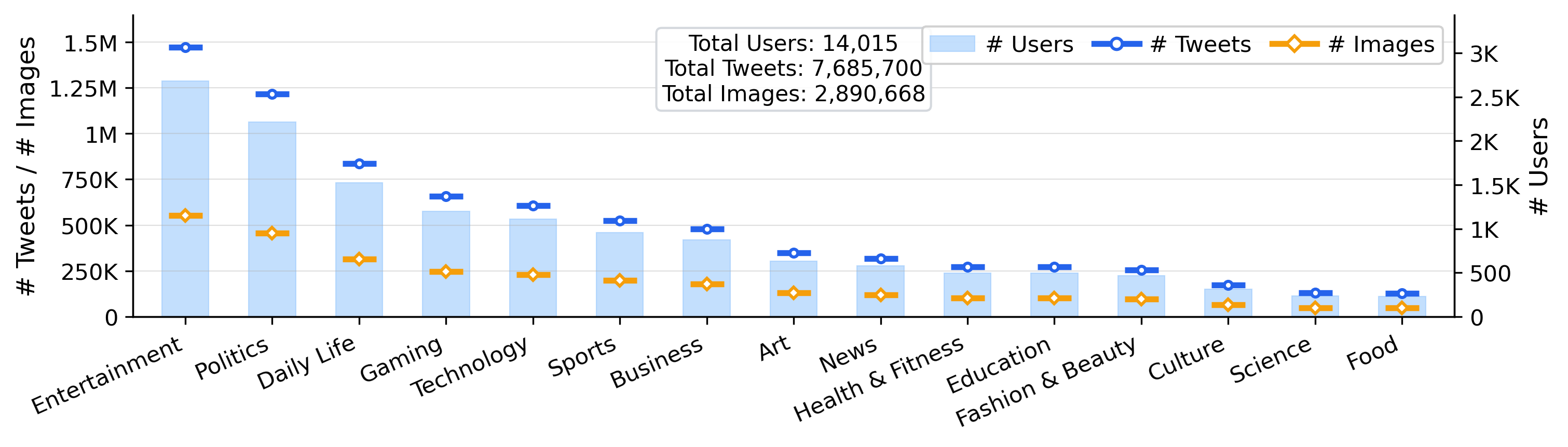}
    \vspace{-1.5em}
    \caption{Data distribution across different interest domains.}
    \label{fig:data-dis}
    \vspace{-2.0em}
\end{figure}

\subsection{Dataset Construction}\label{sec:dataset}
\label{app:dataset construction}
We construct a large-scale multimodal social interaction dataset from X (formerly twitter) to support drift-aware user modeling. As shown in Fig.~\ref{fig:data-dis}, the dataset includes user profiles, interaction behaviors (posts and reposts), multimodal content, and timestamps, covering 15 high-level interest domains with diverse topical distributions. For each domain, active users are first identified followed by collecting their historical interactions, including textual posts, associated images, and profile metadata.
Overall, the dataset comprises 14,015 users, 7,685,700 posts, and 2,890,668 images.

Based on the collected dataset, we construct training samples via a three-step procedure:
{(1) Temporal segmentation.}
For each user, interactions are chronologically ordered and partitioned into non-overlapping collections, each containing 30 posts.
{(2) Pseudo-label annotation.}
For each target collection, we assign an interest label from 15 predefined domains using GPT-5.1. The annotation is conditioned on the content of the target collection and up to 2 preceding collections as historical context. This design incorporates temporal dependencies to better capture interest drift and improve label consistency.
{(3) Quality verification.}
To assess annotation quality, we manually inspect a randomly sampled subset of labeled collections. The results indicate that the generated labels are largely consistent with human judgments.



\subsection{Experimental settings}

\paragraph{Baselines.}
We compare our method with representative baselines from three categories, covering key paradigms in user modeling: sequential recommendation, multi-interest modeling, and content-aware representation learning. 
\begin{itemize}[leftmargin=1.5em, itemindent=0.8em]
    \item \textit{Sequential recommendation.}
We include three representative baselines for modeling user behavior from interaction sequences. SASRec~\cite{kang2018self} uses causal self-attention to model sequential dependencies. BERT4Rec~\cite{sun2019bert4rec} employs a masked language modeling objective for bidirectional sequence modeling. PTUM~\cite{wu2020ptum} incorporates self-supervised learning signals, including masked behavior and next-item prediction.

\item \textit{Multi-interest modeling.}
We include two representative baselines that explicitly model multiple user interests. MIND~\cite{li2019multi} employs capsule-based dynamic routing to extract latent interest representations from behavior sequences. HORAE~\cite{hu2025horae} combines temporal-aware pre-training with explicit interest decomposition to  capture temporal dynamics and multi-interest structures.
HM4SR~\cite{zhang2025hierarchical} integrates hierarchical temporal modeling with a mixture-of-experts architecture to capture evolving user interests from multimodal sequential behaviors.

\item \textit{Content-aware representation learning.}
We include two representative baselines that incorporate auxiliary content information into user modeling. PeterRec~\cite{yuan2020parameter} adopts a pre-training and fine-tuning paradigm to learn transferable user representations from behavioral data. UniSRec~\cite{hou2022towards} integrates textual content with Mixture-of-Experts adapters to learn unified sequential representations across domains.
\end{itemize}

\paragraph{Metrics.}
For \underline{user interest classification}, we evaluate performance using macro Recall, Hit@1, NDCG@3, and KL divergence. Hit@1 reflects top-1 prediction accuracy, while NDCG@3 captures the quality of ranked predictions with higher emphasis on top positions. KL divergence measures the alignment between the predicted distribution and the label-smoothed ground-truth distribution. For \underline{interaction prediction}, we report Accuracy (Acc), Recall, F1 score, and binary cross-entropy (BCE), providing both classification-based and probabilistic evaluation of model performance.

\paragraph{Implementation details.}
The backbone network is trained using the proposed three-stage strategy with a batch size of 128. After the pre-training stage, its parameters are frozen for subsequent optimization. The dataset is split chronologically at the user-collection level, with each user's 12 collections allocated as 8 for training, 2 for validation, and 2 for testing. All experiments use three random seeds. Mean results are reported in main text, with full statistics provided in Appendix~\ref{app:statistical_results}.

\subsection{Experiment I: User Interest Classification}


This experiment aims to evaluate the quality of learned user representations by measuring how accurately they predict users' interest categories. We implement all baselines using their publicly available code or official reproductions, following the same data splits and evaluation protocol for fair comparison.

\begin{table}[htbp]
\centering
\small
\caption{Interest classification performance. $\uparrow/\downarrow$ indicates higher/lower is better.}
\begin{tabular}{lcccc}
\toprule
\textbf{Method} 
& \textbf{Hit@1} $\uparrow$ 
& \textbf{NDCG@3} $\uparrow$ 
& \textbf{Recall} $\uparrow$ 
& \textbf{KL Divergence} $\downarrow$ \\
\midrule

SASRec
& 0.767
& 0.872
& 0.695
& 0.609 \\

BERT4Rec
& 0.769
& 0.874
& 0.699
& 0.586 \\

PTUM
& \cellcolor{blue!10}0.808
& \cellcolor{blue!10}0.899
& \cellcolor{blue!10}0.751
& \cellcolor{blue!10}0.487 \\

MIND
& \cellcolor{blue!25}0.816
& \cellcolor{blue!25}0.903
& \cellcolor{blue!25}0.759
& \cellcolor{blue!25}0.478 \\

HORAE
& 0.801
& 0.896
& 0.740
& 0.512 \\

HM4SR
& 0.759
& 0.866
& 0.688
& 0.615 \\

PeterRec
& 0.785
& 0.886
& 0.717
& 0.537 \\

UniSRec
& 0.763
& 0.869
& 0.681
& 0.609 \\

\midrule

\oursName
& \cellcolor{blue!40}{\textbf{0.874}}
& \cellcolor{blue!40}{\textbf{0.941}}
& \cellcolor{blue!40}{\textbf{0.839}}
& \cellcolor{blue!40}{\textbf{0.318}} \\

\bottomrule
\end{tabular}
\label{tab:classification}
\end{table}

As shown in Table~\ref{tab:classification}, \oursName\ consistently outperforms all baselines across all evaluation metrics. It achieves a Hit@1 of \textbf{0.874}, a Recall of \textbf{0.839}, and an NDCG@3 of \textbf{0.941}, demonstrating its effectiveness in accurately identifying the most relevant target while maintaining strong ranking quality. The substantial improvement in Recall further indicates that our method captures a broader set of relevant candidates, which is critical in recommendation scenarios with diverse user preferences. Compared to the strongest baselines MIND and PTUM, our method improves Hit@1 by approximately \textbf{7.1\%} and \textbf{8.1\%}, respectively, and yields a notable gain in Recall (\textbf{+10.5\%} over MIND). Relative to HORAE, which explicitly models multi-interest representations, our approach achieves an \textbf{9.1\%} improvement in Hit@1, highlighting its advantage in learning more precise and comprehensive user representations. In contrast, purely sequential models such as SASRec and BERT4Rec lag significantly behind (e.g., \textbf{12.2\%} lower Hit@1 for SASRec), suggesting that sequence-only modeling without explicit disentanglement struggles to capture complex user behaviors.

Moreover, \oursName\ achieves the lowest KL divergence of \textbf{0.318}, substantially below all baselines and corresponding to a \textbf{33.5\%} reduction over the second-best method MIND (0.478), indicating significantly better calibration of the predicted interest distribution. Most competing methods exhibit KL values in the range of 0.478--0.615, reflecting notable distributional mismatch even when ranking performance is relatively strong. The superior NDCG@3 further confirms that our method not only identifies the correct target but also ranks relevant items more effectively at top positions. Collectively, these results demonstrate that our framework achieves both strong ranking performance and improved distributional alignment, validating the effectiveness of the proposed representation learning strategy.

\subsection{Experiment II: User Interaction Prediction}
This experiment evaluates the model's ability to predict future user interactions. Given a chronological sequence of behavioral windows for a user, we use the representation derived from our method to predict items in the subsequent window. We adopt the same lightweight prediction head and training protocol for all baselines to ensure fair comparison. Further details are provided in Appendix~\ref{app:impl:downstream}.

Table~\ref{tab:interaction} summarizes the results. \oursName\ achieves the best performance in Recall, obtaining a score of \textbf{0.922}, clearly outperforming all baselines. In particular, it surpasses the strongest baseline HORAE (Recall: 0.900) by approximately \textbf{2.4\%}, demonstrating its stronger capability in retrieving future positive interactions. In terms of F1, \oursName\ achieves \textbf{0.888}, ranking second among all methods and remaining highly competitive with the best-performing baseline PTUM (0.889). For Accuracy, \oursName\ achieves \textbf{0.883}, which is slightly lower than PTUM (0.888), MIND (0.886), and PeterRec (0.884), but remains competitive with these methods. For BCE (lower is better), our method achieves \textbf{0.276}, matching PeterRec (0.276), although it is slightly higher than the best-performing baseline PTUM (0.270). These results demonstrate that our temporal dynamics-aware backbone combined with the MoE interest adapter produces effective user representations for predicting future interactions, with particularly strong advantages in retrieving future positive interactions while maintaining competitive classification performance.


\begin{table}[htbp]
\centering
\small
\caption{Interaction prediction performance. $\uparrow/\downarrow$ indicates higher/lower is better.}
\begin{tabular}{lcccc}
\toprule
\textbf{Method} 
& \textbf{Acc} $\uparrow$ 
& \textbf{Recall} $\uparrow$ 
& \textbf{F1} $\uparrow$ 
& \textbf{BCE} $\downarrow$ \\
\midrule

SASRec
& 0.875
& 0.889
& 0.876
& 0.295 \\

BERT4Rec
& 0.878
& 0.892
& 0.880
& 0.292 \\

PTUM
& \cellcolor{blue!40}0.888
& 0.898
& \cellcolor{blue!40}0.889
& \cellcolor{blue!40}0.270 \\

MIND
& \cellcolor{blue!25}0.886
& 0.897
& \cellcolor{blue!10}0.887
& \cellcolor{blue!25}0.273 \\

HORAE
& 0.884
& \cellcolor{blue!25}0.900
& 0.886
& 0.276 \\

HM4SR
& 0.867
& 0.878
& 0.869
& 0.307 \\

PeterRec
& \cellcolor{blue!10}0.884
& \cellcolor{blue!10}0.900
& 0.886
& \cellcolor{blue!10}0.276 \\

UniSRec
& 0.875
& 0.894
& 0.878
& 0.295 \\

\midrule

\textbf{\oursName}
& \textbf{0.883}
& \cellcolor{blue!40}{\textbf{0.922}}
& \cellcolor{blue!25}{\textbf{0.888}}
& \cellcolor{blue!10}{\textbf{0.276}} \\

\bottomrule
\end{tabular}
\label{tab:interaction}
\end{table}

\subsection{Ablation Study}

We perform ablation experiments on downstream tasks using the same evaluation protocol (Table~\ref{tab:ablation}). 
We study the effect of key components, including (1) backbone network, (2) mixture-of-experts interest adapter, and (3) the training strategy and objective.

\paragraph{Backbone Network.}
The backbone ablations (Static Only, Short-term Only, and Long-term Only) reveal that each component contributes meaningfully to the final performance. Among single-branch variants, Long-term Only achieves the highest Hit@1 (0.863), but still trails the full model by 1.1\%, while Short-term Only (0.809) and Static Only (0.781) lag further behind. To further evaluate the contribution of temporal modeling, we also compare against a content-only baseline (UNITE Mean-Pool), which achieves lower Hit@1 (0.828) and interaction Recall (0.877) than the full model (0.874 and 0.922), with higher KL (0.436) and BCE (0.333). Overall, these results demonstrate the importance of jointly integrating complementary static and multi-scale temporal information with pretrained content representations to learn more comprehensive and effective user representations.

\begin{table}[h]
\centering
\small
\caption{Ablation study on downstream tasks. $\uparrow$/$\downarrow$ indicates higher/lower is better.}
\label{tab:ablation}
\setlength{\tabcolsep}{4pt}
\begin{tabular}{lcccccccc}
\toprule
 & \multicolumn{4}{c}{\textbf{Interest Classification}} & \multicolumn{4}{c}{\textbf{Interaction Prediction}} \\
\cmidrule(lr){2-5} \cmidrule(lr){6-9}
\textbf{Variant} & Hit@1 $\uparrow$ & NDCG@3 $\uparrow$ & Recall $\uparrow$ & KL $\downarrow$ & Acc $\uparrow$ & Recall $\uparrow$ & F1 $\uparrow$ & BCE $\downarrow$ \\
\midrule

Static Only
& 0.781 & 0.880 & 0.741 & 0.560
& 0.809 & 0.839 & 0.814 & 0.409 \\

Short-term Only
& 0.809 & 0.898 & 0.743 & 0.464
& 0.846 & 0.852 & 0.847 & 0.350 \\

Long-term Only
& 0.863 & 0.936 & 0.838 & 0.340
& 0.862 & 0.892 & 0.866 & 0.319 \\

UNITE Mean-Pool
& 0.828 & 0.915 & 0.770 & 0.436
& 0.856 & 0.877 & 0.859 & 0.333 \\

\midrule

w/o Adapter
& \cellcolor{blue!25}0.872 & \cellcolor{blue!25}0.940 & 0.837 & 0.321
& 0.862 & \cellcolor{blue!10}0.885 & 0.865 & 0.319 \\

w/o Gating
& \cellcolor{blue!25}0.872 & \cellcolor{blue!25}0.940 & \cellcolor{blue!10}0.839 & \cellcolor{blue!40}0.316
& 0.864 & 0.883 & 0.867 & 0.316 \\

\midrule

w/o Stage-wise Training
& 0.863 & 0.935 & \cellcolor{blue!40}0.848 & 0.343
& \cellcolor{blue!25}0.867 & \cellcolor{blue!25}0.887 & \cellcolor{blue!25}0.870 & \cellcolor{blue!25}0.309 \\

$\mathcal{L}_\text{cls}$ Only
& 0.868 & 0.937 & \cellcolor{blue!25}0.844 & \cellcolor{blue!40}0.316
& 0.795 & 0.847 & 0.805 & 0.441 \\

$\mathcal{L}_\text{rec}$ Only
& 0.805 & 0.901 & 0.736 & 0.469
& \cellcolor{blue!25}0.867 & 0.882 & \cellcolor{blue!10}0.869 & \cellcolor{blue!10}0.310 \\

\midrule

\textbf{Full model}
& \cellcolor{blue!40}{\textbf{0.874}} & \cellcolor{blue!40}{\textbf{0.941}} & \cellcolor{blue!10}\textbf{0.839} & \cellcolor{blue!10}{\textbf{0.318}}
& \cellcolor{blue!40}{\textbf{0.883}} & \cellcolor{blue!40}{\textbf{0.922}} & \cellcolor{blue!40}{\textbf{0.888}} & \cellcolor{blue!40}{\textbf{0.276}} \\

\bottomrule
\end{tabular}
\end{table}

\paragraph{Mixture-of-Experts Interest Adapter.}
Removing the adapter pool (w/o Adapter) and replacing sparse gating with uniform averaging (w/o Gating) both degrade performance compared with the full model. Specifically, w/o Adapter achieves lower Hit@1 (0.872) and higher BCE (0.319), while w/o Gating yields Hit@1 of 0.872 and BCE of 0.316. Both variants also underperform the full model on interaction prediction (Acc: 0.862 and 0.864 vs.\ 0.883), validating that both the adapter pool and adaptive routing are important for capturing diverse user interests.

\paragraph{Training Strategy and Objective.}
Disabling the stage-wise training strategy (w/o Stage-wise Training) leads to lower Hit@1 (0.863) and higher KL (0.343) than the full model, although it achieves a higher classification Recall (0.848 vs.\ 0.839). It also degrades interaction prediction (Acc: 0.867, Recall: 0.887, F1: 0.870, BCE: 0.309), confirming the effectiveness of our three-stage approach in coordinating the learning dynamics and improving overall performance. Regarding the training objective, $\mathcal{L}_\text{cls}$ alone drives discriminative interest identification (Hit@1: 0.868, KL: 0.316) but substantially degrades interaction prediction (Acc: 0.795, BCE: 0.441); conversely, $\mathcal{L}_\text{rec}$ alone yields competitive interaction prediction (Acc: 0.867, BCE: 0.310) but underperforms on interest classification (Hit@1: 0.805, KL: 0.469), as reconstruction supervision alone lacks the discriminative signal for accurate interest identification.

\subsection{Expert Behavior Analysis}

To further investigate how the MoE interest adapter captures diverse user interests, we analyze expert specialization and adaptive routing patterns across different interest domains. We provide detailed expert-category correspondence matrices and domain-wise routing statistics in the Appendix~\ref{app:expert_analysis}.

\paragraph{Expert Specialization.}
We first examine expert utilization and its correspondence with users' dominant interest categories. The results reveal a non-uniform but well-distributed utilization pattern, with expert usage generally reflecting the distribution of interest categories. Entertainment and Politics exhibit the highest mean gating values (MG), at 0.197 and 0.194, respectively. The Top Expert column further shows that most categories are primarily associated with their corresponding experts, with Sports and Fashion \& Beauty being primarily associated with the Food and Science experts, respectively. Regarding expert specialization, the Politics expert achieves the highest recall and precision (\textbf{88.80\%} and \textbf{93.50\%}), followed by the Entertainment expert (\textbf{82.20\%} and \textbf{85.10\%}). By comparison, the Science expert exhibits a lower recall of 36.50\%, highlighting differences in specialization across domains. These results indicate that the MoE adapter learns meaningful domain-related representations, while the non-uniform routing patterns also reflect varying degrees of overlap among users' interests.

\paragraph{Adaptive Routing.}
We further analyze the routing entropy across different interest domains to examine how the gating network distributes users across experts. The average routing entropy is \textbf{0.584}, with substantial variations across domains. Politics exhibits the lowest entropy of \textbf{0.269}, indicating more concentrated expert routing, whereas Science and Fashion \& Beauty exhibit higher entropy values of \textbf{1.066} and \textbf{1.049}, respectively, suggesting more distributed expert activation. These differences indicate that the gating network learns domain-dependent routing patterns rather than applying uniform expert assignments to all users. In particular, the contrast between concentrated routing in Politics and more distributed routing in Science and Fashion \& Beauty reflects the adaptive nature of expert selection across different user interest domains.

\subsection{Discussion}
\label{app:disscussion}
While our framework demonstrates strong performance on user interest classification and interaction prediction, several limitations warrant discussion. First, pseudo-label generation relies on GPT-5.1 with contextual signals derived from adjacent behavioral windows. Although this design improves temporal consistency, label quality may still be affected by annotation noise or domain-specific biases inherent to the language model. Future work could explore semi-supervised or self-supervised alternatives to reduce dependence on external supervision. Second, the multi-scale backbone adopts fixed window lengths for short-term (LSTM) and long-term (Transformer) modeling. In practice, optimal temporal granularity likely varies across users depending on their activity patterns and engagement frequency. Adaptive window selection mechanisms could provide finer-grained personalization in such heterogeneous settings. Third, the MoE interest adapter employs a fixed number of experts with static top-$K'$ routing. While effective, the optimal number of activated experts may differ substantially across users with varying behavioral diversity. Dynamic expert allocation conditioned on user-level complexity is a promising direction for further improving adaptability. 
Finally, the current framework operates in an offline training regime. Although our three-stage training strategy promotes stable optimization, online continual learning with efficient incremental updates remains an open challenge. Incorporating lightweight adaptation mechanisms to handle real-time interest drift would be valuable for production-scale deployment. 
Addressing these limitations in future work can further enhance the robustness, efficiency, and generalizability of dynamic user profiling.

\section{Conclusion}
In this paper, we proposed a dynamic user profiling framework that jointly models explicit and implicit preferences across multiple temporal scales. Our approach features a multi-scale backbone fusing static, short-term, and long-term features, a Mixture-of-Experts interest adapter with sparse routing for multi-interest decomposition, and a three-stage training strategy that ensures stable specialization. We also contributed a large-scale multimodal dataset from X with temporally evolving interest labels. 
Extensive experiments on user interest classification and interaction prediction demonstrated that our method consistently achieves strong performance across both tasks.
Ablation studies validated the contribution of each key component. Future work includes extending the framework to more fine-grained temporal dynamics and exploring online adaptation scenarios. Our code and dataset will be made publicly available under the privacy policy to facilitate reproducibility.

\section*{Acknowledgments}
Nan Cao is the corresponding author. This work was supported by the National Key Research and Development Program of China (2023YFB3107100).

\newpage

\bibliographystyle{unsrt}
\bibliography{references}

\newpage

\appendix

\section{Qualitative Analysis}
\label{app:qualitative}
\subsection{Interest Drift Case Study}
\label{app:qualitative:drift}

\begin{figure}[h]
    \centering
    \begin{subfigure}{\textwidth}
        \includegraphics[width=\textwidth]{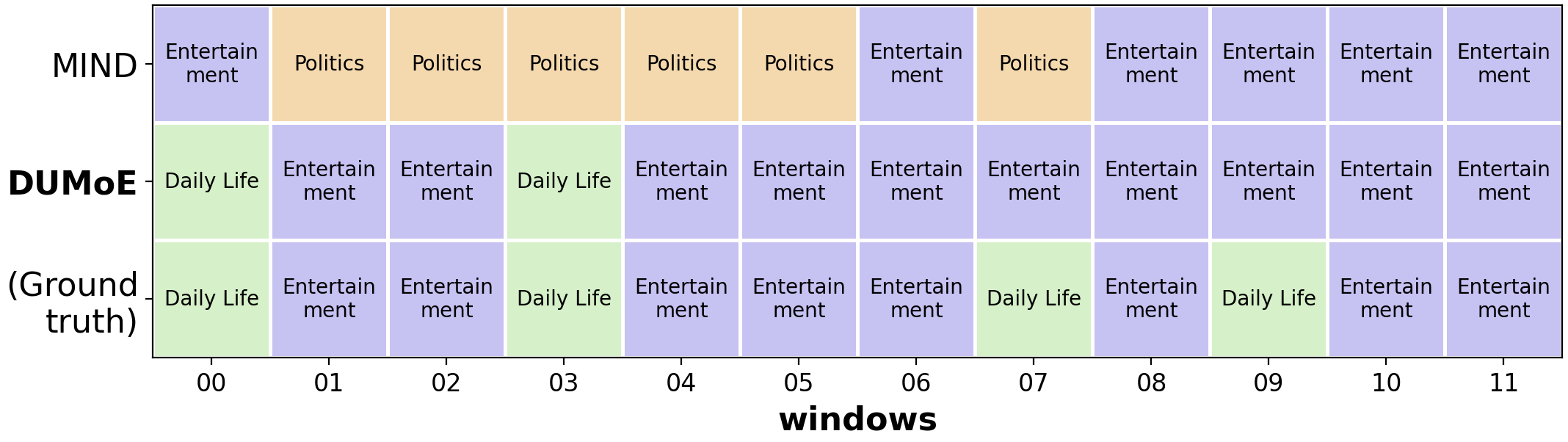}
        \caption{User A: Entertainment-dominant interest with occasional drift to Daily Life.}
    \end{subfigure}
    \vspace{0.8em}
    \begin{subfigure}{\textwidth}
        \includegraphics[width=\textwidth]{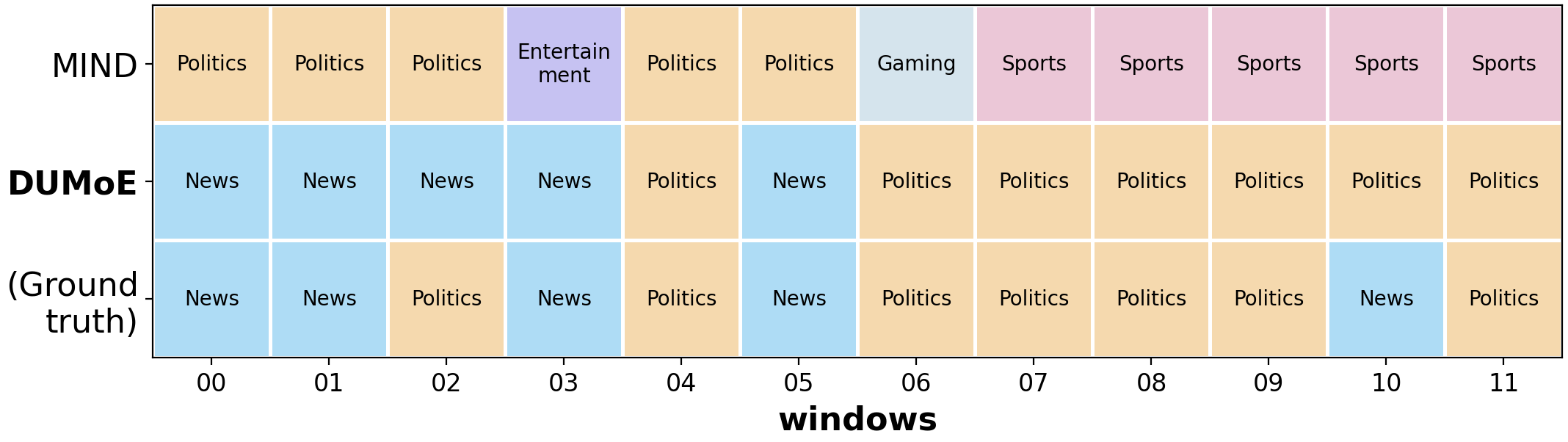}
        \caption{User B: Alternating News--Politics interest.}
    \end{subfigure}
    \vspace{0.8em}
    \begin{subfigure}{\textwidth}
        \includegraphics[width=\textwidth]{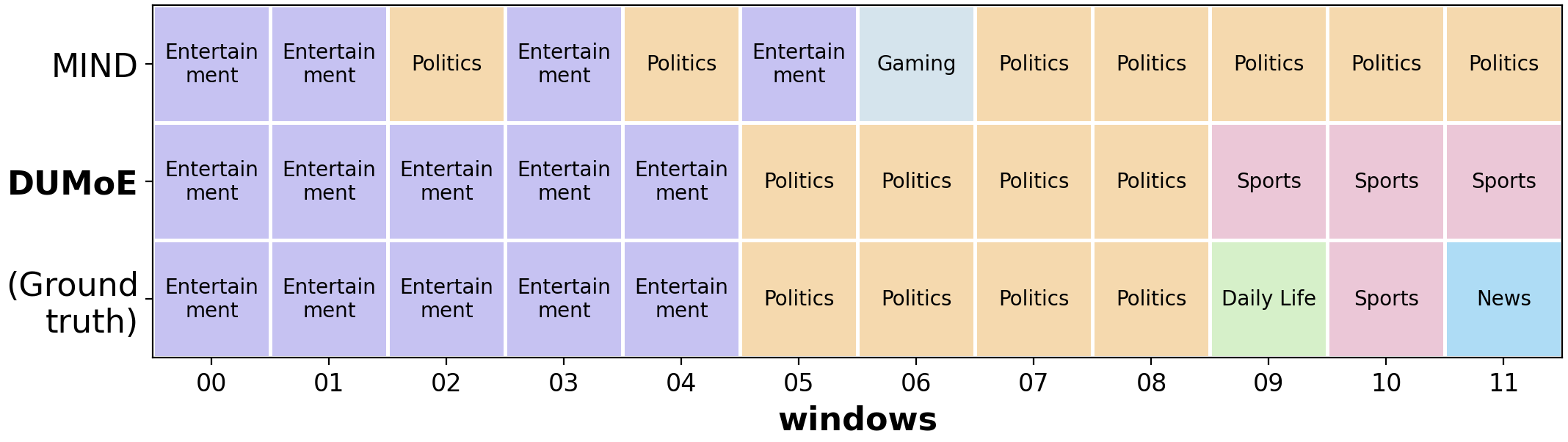}
        \caption{User C: Entertainment-to-Politics shift followed by late-stage drift.}
    \end{subfigure}
    \caption{Window-level interest trajectories for three representative users.}
    \label{fig:case_study}
\end{figure}

Figure~\ref{fig:case_study} visualizes window-level interest predictions from MIND and DUMoE alongside ground-truth labels for three representative users. We highlight User B as an illustrative case. The ground truth shows an alternating \textit{News}--\textit{Politics} pattern across 12 windows, reflecting fine-grained interest fluctuation. MIND fails to track this, instead drifting into unrelated categories such as \textit{Gaming} and \textit{Sports} in later windows. DUMoE consistently captures the \textit{News}--\textit{Politics} alternation, with only minor deviations, demonstrating its ability to model co-existing interests and short-term drift. The other two users exhibit similar trends: DUMoE more faithfully follows ground-truth transitions while MIND tends to collapse to a dominant category or produce inconsistent predictions. 

\section{Broader Impact}
\label{app:impact}
This work advances the ability of machine learning systems to model temporally evolving user interests from noisy social media data, with potential benefits for recommendation, content filtering, and personalized information access. The sparse Mixture-of-Experts design further promotes interpretability by routing users to disentangled interest subspaces rather than opaque monolithic representations. We also release a large-scale multimodal dataset to support future research in user modeling and social computing; to protect privacy, only anonymized user IDs and tweet IDs are released, with content filtering applied prior to distribution.

We acknowledge that fine-grained user interest modeling carries inherent dual-use risks. Temporal user representations could be misused for surveillance, behavioral profiling, or the targeted delivery of manipulative content timed to moments of heightened susceptibility. Closely tracking interest drift may also deepen filter bubbles by reinforcing existing preferences at the expense of viewpoint diversity; we recommend that systems built on our representations incorporate diversity constraints as a complementary objective. Additionally, our pseudo-label pipeline relies on an LLM, whose training biases may produce less accurate annotations for underrepresented or non-English-speaking communities, potentially leading to uneven model performance across demographic groups. To mitigate these risks, we distribute the dataset under terms of use that prohibit surveillance and targeted manipulation, and encourage practitioners to engage with ethics review processes when deploying user behavior modeling in high-stakes contexts.

\section{Dataset Details}
\label{app:dataset}

\subsection{Dataset Statistics}
\label{app:dataset:stats}

Table~\ref{tab:domain_stats} reports per-domain statistics of our dataset.
The data were collected from X (formerly Twitter), and the interest taxonomy is derived by consolidating platform-specific community classifications into \textbf{15 high-level interest domains}. 
Overall, the dataset comprises \textbf{14,015} users, \textbf{7,685,700} tweets, and \textbf{2,890,668} associated images.
The pronounced skew across domains reflects the organic distribution of user interests on the platform and motivates the class-weighted focal loss adopted in our learning objective (Eq.~15 of the main paper).

\begin{table}[h]
\centering
\small
\caption{Per-domain dataset statistics.}
\label{tab:domain_stats}
\setlength{\tabcolsep}{8pt}
\begin{tabular}{lrrr}
\toprule
\textbf{Domain} & \textbf{\# Users} & \textbf{\# Tweets} & \textbf{\# Images} \\
\midrule
Entertainment    & 2,681 & 1,470,236 & 552,971 \\
Politics         & 2,217 & 1,215,783 & 457,268 \\
Daily Life       & 1,524 &   835,748 & 314,333 \\
Gaming           & 1,201 &   658,618 & 247,713 \\
Technology       & 1,108 &   607,617 & 228,531 \\
Sports           &   957 &   524,810 & 197,386 \\
Business         &   872 &   478,197 & 179,855 \\
News             &   580 &   318,067 & 119,628 \\
Health \& Fitness &   499 &   273,647 & 102,921 \\
Education        &   494 &   270,905 & 101,890 \\
Fashion \& Beauty &   466 &   255,550 &  96,115 \\
Art              &   634 &   347,680 & 130,766 \\
Culture          &   314 &   172,195 &  64,764 \\
Science          &   236 &   129,420 &  48,676 \\
Food             &   232 &   127,227 &  47,851 \\
\midrule
\textbf{Total}   & \textbf{14,015} & \textbf{7,685,700} & \textbf{2,890,668} \\
\bottomrule
\end{tabular}
\end{table}

\subsection{Pseudo-Label Annotation Protocol}
\label{app:dataset:annotation}

To obtain large-scale temporally-evolving interest labels, we use GPT-5.1 to automatically annotate each user's tweet sequence. For each user, tweets are first sorted in descending chronological order and partitioned into non-overlapping groups of 30, yielding $G_u$ ordered snapshots. We then construct labeling units using a sliding-window scheme: each window $t$ (with $t = 0, 1, \ldots, G_u - 1$) consists of up to three consecutive groups $(t, t{+}1, t{+}2)$, forming a short-term context of at most 90 tweets. This design allows each group to appear in multiple overlapping windows, enriching temporal coverage. Due to API constraints, each window contains at most 50 images.

Each window is converted into a multimodal prompt with three sub-windows (A, B, C), interleaving tweet text and associated images. The prompt used in our implementation is as follows:

\begin{tcolorbox}[
    colback=gray!10,
    colframe=black,
    title=Multimodal Interest Classification Prompt,
    fonttitle=\bfseries,
    breakable
]

You are a precise multimodal content classification assistant. You analyze both text and images jointly to infer user interest. Follow category definitions strictly and avoid introducing external assumptions.

\medskip
Please analyze the user's tweets across three temporal windows and determine the single most relevant interest category.

\begin{enumerate}[leftmargin=*,topsep=2pt,itemsep=2pt]
    \item Window A (Group $t$)
    \item Window B (Group $t{+}1$)
    \item Window C (Group $t{+}2$)
\end{enumerate}

Available categories (15 in total):  
Entertainment, Sports, Technology, Business, Education, Gaming, Politics, Art, Culture, Science, Food, Daily Life, Fashion \& Beauty, Health \& Fitness, News.

\medskip
Each window contains interleaved tweet text and associated images.

\medskip
\textbf{Output Requirement:}  
Return ONLY the single best-matching category name. Do not output explanations, punctuation, or additional text.

\end{tcolorbox}

We evaluated multiple large language model families, including GPT, Claude, and Gemini, on a pilot set of 8 users in terms of accuracy, stability, and cost. GPT-based models consistently demonstrated the best overall trade-off, combining high classification accuracy with stable outputs and low refusal rates on multimodal inputs. Within the GPT family, we further compared GPT-5.1, GPT-5.2, and GPT-5.4. While GPT-5.4 achieves slightly higher accuracy (93.2\%), it is significantly slower and more expensive; GPT-5.1 achieves comparable performance (92.1\%) with better stability and fewer refusals, and is thus selected (temperature $= 0.3$, max tokens $= 20$). 
Finally, we randomly sampled approximately 5\% of users ($\approx$700) for manual verification. Three independent annotators with domain expertise in social media content analysis labeled the same sampled windows using the 15-category taxonomy. The average pairwise inter-annotator agreement was \textbf{85.45\%}, with a Fleiss' kappa of \textbf{0.838}, indicating strong consistency among human annotators. The agreement between GPT-5.1 annotations and human judgments reached \textbf{92.3\%}, further supporting the reliability of our pseudo-label annotation pipeline.

\section{Implementation and Reproducibility}
\label{app:impl}

All experiments are conducted on a server equipped with 2$\times$ NVIDIA A800 80GB PCIe GPUs (each with 80\,GB memory). The software environment consists of PyTorch 2.10.0+cu128 and CUDA 12.8. Mixed-precision training (AMP, \texttt{float16}) is enabled throughout to reduce memory usage and accelerate training.

\subsection{Input Encoding and Preprocessing}
\label{app:impl:arch}

\paragraph{Interaction Sequence Construction.}
Following the dataset construction protocol (Section~\ref{sec:dataset}), each user's interaction history is partitioned into non-overlapping groups of 30 posts in chronological order. The short-term sequence $\mathcal{S}^t_u$ covers the most recent $L_t = 3$ groups (90 posts), while the long-term sequence $\mathcal{S}^l_u$ spans the most recent $L_l = 10$ groups (300 posts), such that $\mathcal{S}^t_u \subset \mathcal{S}^l_u$. Within each window, post embeddings are first mean-pooled within each group to obtain one group-level embedding, yielding sequences of length $L_t$ and $L_l$ respectively. These group-level sequences are then fed into $\phi_t(\cdot)$ (LSTM) and $\phi_l(\cdot)$ (Transformer) as described in Section~\ref{sec:backbone}, with $\mathbf{r}^t_u$ taken as the final hidden state of the LSTM and $\mathbf{r}^l_u$ taken as the \texttt{[CLS]} token output of the Transformer. Users with fewer than $L_l$ groups in their history use all available groups, with zero-padding applied as needed. The window lengths $L_t = 3$ and $L_l = 10$ are chosen to maintain a meaningful separation between temporal scales: the short-term window captures recent behavioral shifts, while the long-term window provides broader historical context sufficient to model stable interest patterns.

\paragraph{Post and Profile Encoding.}
We use UNITE-Base-Qwen2-VL-2B~\cite{kong2025modality} (1536-d output) as the frozen multimodal encoder for all text and image inputs. Each post embedding is obtained by jointly encoding the tweet text and associated image. All embeddings are pre-extracted offline prior to training.
We construct a 1542-d representation $\mathbf{a}_u$ by concatenating textual and numerical features as detailed in Table~\ref{tab:profile_encoding}, which is then fed into the static MLP $\phi_s(\cdot)$.

\begin{table}[h]
\centering\small
\caption{User profile attribute encoding.}
\label{tab:profile_encoding}
\begin{tabular}{lll}
\toprule
\textbf{Field} & \textbf{Processing} & \textbf{Dim} \\
\midrule
\texttt{screen\_name}, \texttt{name}, \texttt{description}, \texttt{location} & UNITE encode $\rightarrow$ mean & 1536 \\
\texttt{verified}         & binarize                 & 1 \\
\texttt{media\_count}     & binarize + log-scale     & 2 \\
\texttt{followers\_count} & log-scale to $[0,1]$     & 1 \\
\texttt{friends\_count}   & log-scale to $[0,1]$     & 1 \\
\texttt{statuses\_count}  & log-scale to $[0,1]$     & 1 \\
\midrule
\multicolumn{2}{l}{Concatenation} & 1542 \\
\bottomrule
\end{tabular}
\end{table}

\subsection{Three-Stage Training Details}
\label{app:impl:training}

We optimize the backbone and MoE adapter with AdamW (weight decay $10^{-2}$, gradient clipping 1.0) using a three-stage schedule, with all stages using a batch size of 128. In Stage 1 (backbone pre-training, lr $1\times10^{-4}$, 15 epochs), the backbone is trained together with prediction heads $\phi_{\mathrm{cls}}$ and $\phi_{\mathrm{rec}}$ under the two-task objective balanced by $\lambda=1.5$, while all adapter and gating parameters are frozen. In Stage 2 (category-guided expert specialization, lr $5\times10^{-4}$, 5 epochs), the backbone is frozen, and each of the $K=15$ experts is trained using only samples whose dominant interest label matches its corresponding category. Specifically, expert $k$ is optimized on samples with label $y=k$, while the recommendation head is jointly optimized alongside each expert. This category-guided training strategy encourages each expert to capture category-related user representations. In Stage 3 (gating optimization, lr $1\times10^{-3}$, 5 epochs), both the backbone and experts are frozen, and only the gating parameters $(W_g, \mathbf{b}_g)$ are updated with sparse top-$K'=3$ routing.

The model architecture is configured as follows: the short-term encoder is a 1-layer LSTM with hidden size 256; the long-term encoder is a 2-layer Transformer with 8 attention heads and FFN dimension 1024; each expert adapter is a bottleneck MLP with 64-d hidden size and GELU activation; the fusion head $\phi_f$ projects to 256-d with dropout 0.4.

\subsection{KL-Based Gating Mechanism}
\label{app:kl_gating}

At inference, the backbone representation $\mathbf{z}_u$ and the MoE-adapted representation $\mathbf{h}_u$ are fused via a KL-based gate that uses interest drift intensity as an adaptive signal. To estimate drift, we compute a reference backbone representation $\mathbf{z}^-_u$ by removing the most recent $L_t$ posts from both branches, retaining only the distal portion $[L_t{+}1,\, L_l]$ of the long-term sequence $\mathcal{S}^l_u$. Both $\mathbf{z}_u$ and $\mathbf{z}^-_u$ are produced by the same backbone defined in Eq.~\eqref{eq:fusion} with shared parameters. The KL divergence between their projected interest distributions measures how much the most recent behavioral context shifts the user's interest prediction — a large divergence indicates strong short-term drift, whereas a small divergence suggests that recent posts carry little new information. The gate then adaptively weights $\mathbf{z}_u$ and $\mathbf{h}_u$ based on this signal, as summarized in Algorithm~\ref{alg:kl_gating}.

\begin{algorithm}[h]
\caption{KL-Based Inference Gating}
\label{alg:kl_gating}
\begin{algorithmic}[1]
\Require Backbone repr $\mathbf{z}_u$, short-term-removed backbone repr $\mathbf{z}^-_u$, MoE repr $\mathbf{h}_u$, classifier $\phi_{\mathrm{cls}}$
\Require $s=0.01$, $\alpha_{\min}=0.05$, $\alpha_{\max}=0.80$, $\tau=0.012$
\Ensure Fused representation $\tilde{\mathbf{h}}_u$
\State $\mathbf{p}_u \leftarrow \phi_{\mathrm{cls}}(\mathbf{z}_u)$,\quad $\mathbf{p}^-_u \leftarrow \phi_{\mathrm{cls}}(\mathbf{z}^-_u)$ \Comment{project to interest simplex}
\State $D_u \leftarrow D_{\mathrm{KL}}(\mathbf{p}_u \| \mathbf{p}^-_u)$ \Comment{drift intensity: effect of recent posts on interest dist.}
\If{$D_u \leq \tau$} \Comment{hard switch: negligible drift}
    \State $\tilde{\mathbf{h}}_u \leftarrow \mathbf{z}_u$
\Else \Comment{soft fusion: larger drift $\Rightarrow$ larger $\alpha_u$ $\Rightarrow$ more MoE}
    \State $H_u \leftarrow -\mathbf{p}_u^{\top} \log \mathbf{p}_u \;/\; \log C$ \Comment{normalized entropy of $\mathbf{p}_u$}
    \State $\alpha_u \leftarrow \sigma\!\left(\dfrac{D_u - \tau}{s}\right) \cdot \left(0.5 + 0.5\,H_u\right)$
    \State $\alpha_u \leftarrow \mathrm{clip}(\alpha_u,\, \alpha_{\min},\, \alpha_{\max})$
    \State $\tilde{\mathbf{h}}_u \leftarrow (1 - \alpha_u)\,\mathbf{z}_u + \alpha_u\,\mathbf{h}_u$
\EndIf
\State \Return $\tilde{\mathbf{h}}_u$
\end{algorithmic}
\end{algorithm}

\subsection{Baseline Implementation Details}
\label{app:impl:baselines}

All baselines are pretrained on the same behavioral sequences following their respective original training objectives: next-item prediction for SASRec, PTUM, MIND, PeterRec, UniSRec, HORAE, and HM4SR, and masked item modeling for BERT4Rec. Following the content-aware setting of UniSRec~\cite{hou2022towards}, we initialize the item embedding table from precomputed UNITE vectors and keep it frozen throughout pretraining, replacing the original id-based embeddings with semantic representations; all other architectural and optimization details follow the original papers. For each window, the input sequence contains up to 300 temporally ordered posts; windows with fewer than 2 valid items are discarded. All baselines use a hidden dimension of 256 and export user representations of the same dimensionality, ensuring compatibility with the downstream prediction heads. Baselines are trained for 25 epochs with AdamW (lr $10^{-3}$, weight decay $10^{-2}$, batch size 128) with mixed-precision enabled, with checkpoint selection by minimum validation loss.

\subsection{Downstream Task Settings}
\label{app:impl:downstream}

All baselines and DUMoE are evaluated on the same train/val/test split and user behavioral windows, using the same frozen UNITE tweet embeddings and training configuration for fair comparison.

\paragraph{User interest classification.}
We adopt a frozen-encoder probing protocol: all model parameters are fixed after training, and a lightweight two-layer MLP probe (hidden size 256, ReLU, dropout 0.2) is trained on top of the exported representations using cross-entropy loss with Adam (lr $10^{-3}$, batch size 512) for 20 epochs. Checkpoint selection is based on validation NDCG@3, and test-set metrics are computed in a single forward pass with no further tuning. KL divergence is reported as a calibration diagnostic rather than a training objective, computed against the label-smoothed target:
\begin{equation}
    \mathbf{t} = 0.9\cdot\mathrm{onehot}(y) + \frac{0.1}{C}\cdot\mathbf{1},
\end{equation}
where $C$ is the number of interest categories.

\paragraph{User interaction prediction.}
For each user $u$ and target window $t$, training samples are constructed as triples $(\mathbf{h}_{u,t-1},\ \mathbf{v}^+,\ \mathbf{v}^-)$, where $\mathbf{h}_{u,t-1}$ is the frozen user representation derived from window $t-1$, ensuring no future information leakage. For each $(u, t)$ pair, we sample 10 \textbf{positive posts} $\mathbf{v}^+$ from those the user actually interacted with within window $t$. \textbf{Negative posts} $\mathbf{v}^-$ are sampled at a 1:1 ratio from posts that satisfy: (1) falling within the same time range as window $t$, and (2) not belonging to user $u$. This time-range constraint encourages the model to capture genuine user--post affinity rather than temporal co-occurrence patterns. 
The prediction head first projects the post embedding $\mathbf{f} \in \mathbb{R}^{1536}$ into a 256-d space via a linear layer with ReLU and dropout ($p=0.3$), then concatenates it with the user representation $\mathbf{z}_{u,t-1} \in \mathbb{R}^{256}$ and passes the result through a two-layer MLP to produce a scalar score $s$. The head is trained for 5 epochs with the BCE-based negative sampling loss defined in section ~\ref{app:method 4.3}, with only the projection and scoring parameters updated while the user encoder remains frozen.

\subsection{Full Statistical Results}
\label{app:statistical_results}

We report the full results of user interest classification and interaction prediction over three random seeds (42, 52, and 62). Each entry is presented as the mean and standard deviation across the three runs. For the classification task(Table~\ref{tab:full_interest_results}), DUMoE significantly outperforms all baselines across all metrics (mean p-value < 0.01). For the interaction prediction task(Table~\ref{tab:full_interaction_results}), DUMoE achieves comparable results with the strongest baselines and significantly outperforms SASRec, BERT4Rec, UniSRec, and HM4SR (mean p-value < 0.05).

We also provide the full statistical results of the ablation study in Tables~\ref{tab:ablation_interest_results} and~\ref{tab:ablation_interaction_results}. The results further detail the contributions of individual architectural components and training objectives to both tasks.

\clearpage

\begin{table*}[!p]
\centering
\small
\caption{Full results of user interest classification. Values are mean $\pm$ standard deviation. The last column reports the mean $p$-value of each baseline compared with DUMoE.}
\label{tab:full_interest_results}
\setlength{\tabcolsep}{7pt}
\begin{tabular}{lccccc}
\toprule
\textbf{Model} & \textbf{Hit@1} $\uparrow$ & \textbf{NDCG@3} $\uparrow$ & \textbf{Recall} $\uparrow$ & \textbf{KL} $\downarrow$ & \textbf{Mean $p$} \\
\midrule
SASRec & 0.7669$\pm$0.0147 & 0.8717$\pm$0.0081 & 0.6950$\pm$0.0196 & 0.6092$\pm$0.0180 & $2.87\times10^{-3}$ \\
BERT4Rec & 0.7687$\pm$0.0055 & 0.8744$\pm$0.0038 & 0.6991$\pm$0.0058 & 0.5855$\pm$0.0128 & $3.36\times10^{-4}$ \\
PTUM & 0.8082$\pm$0.0022 & 0.8990$\pm$0.0007 & 0.7513$\pm$0.0040 & 0.4872$\pm$0.0041 & $5.35\times10^{-4}$ \\
MIND & 0.8158$\pm$0.0035 & 0.9034$\pm$0.0005 & 0.7594$\pm$0.0022 & 0.4780$\pm$0.0053 & $4.66\times10^{-4}$ \\
HORAE & 0.8006$\pm$0.0047 & 0.8963$\pm$0.0008 & 0.7403$\pm$0.0042 & 0.5121$\pm$0.0059 & $5.45\times10^{-4}$ \\
HM4SR & 0.7587$\pm$0.0066 & 0.8658$\pm$0.0032 & 0.6884$\pm$0.0091 & 0.6145$\pm$0.0124 & $1.41\times10^{-3}$ \\
PeterRec & 0.7850$\pm$0.0008 & 0.8855$\pm$0.0003 & 0.7172$\pm$0.0037 & 0.5371$\pm$0.0059 & $2.90\times10^{-4}$ \\
UniSRec & 0.7627$\pm$0.0029 & 0.8694$\pm$0.0025 & 0.6808$\pm$0.0057 & 0.6091$\pm$0.0117 & $3.65\times10^{-4}$ \\
\midrule
\textbf{DUMoE} & \textbf{0.8735$\pm$0.0025} & \textbf{0.9410$\pm$0.0019} & \textbf{0.8389$\pm$0.0045} & \textbf{0.3177$\pm$0.0071} & -- \\
\bottomrule
\end{tabular}
\end{table*}

\begin{table*}[!p]
\centering
\small
\caption{Full results of user interaction prediction. Values are mean $\pm$ standard deviation. The last column reports the mean $p$-value of each baseline compared with DUMoE.}
\label{tab:full_interaction_results}
\setlength{\tabcolsep}{7pt}
\begin{tabular}{lccccc}
\toprule
\textbf{Model} & \textbf{Acc} $\uparrow$ & \textbf{Recall} $\uparrow$ & \textbf{F1} $\uparrow$ & \textbf{BCE} $\downarrow$ & \textbf{Mean $p$} \\
\midrule
SASRec & 0.8746$\pm$0.0016 & 0.8890$\pm$0.0093 & 0.8764$\pm$0.0019 & 0.2953$\pm$0.0044 & 0.0216 \\
BERT4Rec & 0.8781$\pm$0.0008 & 0.8920$\pm$0.0104 & 0.8798$\pm$0.0018 & 0.2917$\pm$0.0015 & 0.0176 \\
PTUM & 0.8877$\pm$0.0017 & 0.8979$\pm$0.0098 & 0.8889$\pm$0.0022 & 0.2696$\pm$0.0028 & 0.6610 \\
MIND & 0.8857$\pm$0.0012 & 0.8967$\pm$0.0104 & 0.8869$\pm$0.0022 & 0.2730$\pm$0.0030 & 0.5440 \\
HORAE & 0.8842$\pm$0.0014 & 0.9001$\pm$0.0109 & 0.8860$\pm$0.0024 & 0.2759$\pm$0.0041 & 0.3910 \\
HM4SR & 0.8670$\pm$0.0017 & 0.8783$\pm$0.0057 & 0.8685$\pm$0.0022 & 0.3071$\pm$0.0025 & 0.0023 \\
PeterRec & 0.8844$\pm$0.0011 & 0.8996$\pm$0.0056 & 0.8862$\pm$0.0011 & 0.2755$\pm$0.0023 & 0.3780 \\
UniSRec & 0.8752$\pm$0.0005 & 0.8939$\pm$0.0103 & 0.8775$\pm$0.0014 & 0.2950$\pm$0.0017 & 0.0158 \\
\midrule
\textbf{DUMoE} & \textbf{0.8831$\pm$0.0011} & \textbf{0.9220$\pm$0.0030} & \textbf{0.8875$\pm$0.0009} & \textbf{0.2755$\pm$0.0033} & -- \\
\bottomrule
\end{tabular}
\end{table*}

\begin{table*}[!p]
\centering
\small
\setlength{\tabcolsep}{7pt}
\caption{Full ablation results of user interest classification. Values are mean $\pm$ standard deviation. }
\label{tab:ablation_interest_results}
\begin{tabular}{lcccc}
\toprule
\textbf{Variant} & \textbf{Hit@1} $\uparrow$ & \textbf{NDCG@3} $\uparrow$ & \textbf{Recall} $\uparrow$ & \textbf{KL} $\downarrow$ \\
\midrule
Static Only & 0.7808$\pm$0.0050 & 0.8802$\pm$0.0045 & 0.7409$\pm$0.0099 & 0.5595$\pm$0.0133 \\
Short-term Only & 0.8093$\pm$0.0033 & 0.8980$\pm$0.0030 & 0.7430$\pm$0.0039 & 0.4636$\pm$0.0153 \\
Long-term Only & 0.8632$\pm$0.0042 & 0.9359$\pm$0.0032 & 0.8378$\pm$0.0080 & 0.3402$\pm$0.0116 \\
UNITE Mean-Pool & 0.8283$\pm$0.0042 & 0.9150$\pm$0.0016 & 0.7698$\pm$0.0096 & 0.4358$\pm$0.0078 \\
\midrule
w/o Adapter & 0.8719$\pm$0.0024 & 0.9397$\pm$0.0006 & 0.8365$\pm$0.0054 & 0.3206$\pm$0.0050 \\
w/o Gating & 0.8719$\pm$0.0063 & 0.9395$\pm$0.0043 & 0.8392$\pm$0.0089 & 0.3156$\pm$0.0132 \\
\midrule
w/o Stage-wise Training & 0.8632$\pm$0.0062 & 0.9349$\pm$0.0034 & 0.8476$\pm$0.0059 & 0.3432$\pm$0.0145 \\
$\mathcal{L}_{cls}$ Only & 0.8681$\pm$0.0027 & 0.9373$\pm$0.0012 & 0.8443$\pm$0.0057 & 0.3160$\pm$0.0070 \\
$\mathcal{L}_{rec}$ Only & 0.8045$\pm$0.0118 & 0.9007$\pm$0.0050 & 0.7359$\pm$0.0160 & 0.4694$\pm$0.0109 \\
\midrule

\textbf{Full model} & \textbf{0.8735$\pm$0.0025} & \textbf{0.9410$\pm$0.0019} & \textbf{0.8389$\pm$0.004}5 & \textbf{0.3177$\pm$0.0071} \\
\bottomrule
\end{tabular}
\end{table*}

\begin{table*}[!p]
\centering
\small
\setlength{\tabcolsep}{7pt}
\caption{Full ablation results of user interaction prediction. Values are mean $\pm$ standard deviation.}
\label{tab:ablation_interaction_results}
\begin{tabular}{lcccc}
\toprule
\textbf{Variant} & \textbf{Acc} $\uparrow$ & \textbf{Recall} $\uparrow$ & \textbf{F1} $\uparrow$ & \textbf{BCE} $\downarrow$ \\
\midrule
Static Only & 0.8088$\pm$0.0024 & 0.8393$\pm$0.0211 & 0.8144$\pm$0.0031 & 0.4085$\pm$0.0045 \\
Short-term Only & 0.8464$\pm$0.0006 & 0.8519$\pm$0.0051 & 0.8472$\pm$0.0002 & 0.3499$\pm$0.0010 \\
Long-term Only & 0.8622$\pm$0.0047 & 0.8921$\pm$0.0128 & 0.8662$\pm$0.0024 & 0.3185$\pm$0.0082 \\
UNITE Mean-Pool & 0.8561$\pm$0.0011 & 0.8770$\pm$0.0116 & 0.8590$\pm$0.0013 & 0.3329$\pm$0.0020 \\
\midrule
w/o Adapter & 0.8624$\pm$0.0030 & 0.8849$\pm$0.0182 & 0.8654$\pm$0.0023 & 0.3187$\pm$0.0072 \\
w/o Gating & 0.8641$\pm$0.0044 & 0.8826$\pm$0.0173 & 0.8666$\pm$0.0019 & 0.3155$\pm$0.0077 \\
\midrule
w/o Stage-wise Training & 0.8672$\pm$0.0030 & 0.8867$\pm$0.0079 & 0.8698$\pm$0.0021 & 0.3086$\pm$0.0048 \\
$\mathcal{L}_{cls}$ Only & 0.7948$\pm$0.0037 & 0.8469$\pm$0.0046 & 0.8050$\pm$0.0022 & 0.4409$\pm$0.0040 \\
$\mathcal{L}_{rec}$ Only & 0.8666$\pm$0.0029 & 0.8824$\pm$0.0078 & 0.8687$\pm$0.0018 & 0.3098$\pm$0.0077 \\
\midrule
\textbf{Full model} & \textbf{0.8831$\pm$0.0011} & \textbf{0.9220$\pm$0.0030} & \textbf{0.8875$\pm$0.0009} & \textbf{0.2755$\pm$0.0033} \\
\bottomrule
\end{tabular}
\end{table*}

\clearpage

\section{Expert Analysis}
\label{app:expert_analysis}

The MoE adapter is designed to capture category-related patterns in user representations while accommodating temporal changes in user interests. We further examine its behavior through category-level expert usage statistics and representative cases, focusing on the correspondence between interest categories and expert assignments, routing uncertainty, and adapter activation.

\paragraph{Metrics.}
We use three metrics to characterize expert behavior. First, the mean gate weight (MG) measures the average routing weight assigned to the expert corresponding to category $c$:
\begin{equation}
    \mathrm{MG}_c =
    \frac{1}{N_c}\sum_{u:y_u=c} g_{u,k(c)},
\end{equation}
where $N_c$ is the number of windows with ground-truth label $c$, and $k(c)$ denotes its corresponding expert. Unselected experts are assigned zero gate weight.

Second, mean routing entropy ($\overline{H}_c$) measures the uncertainty of expert selection. For each window $u$, we calculate the entropy of its normalized Top-$K'$ gating weights:
\begin{equation}
    H_u=-\sum_{j\in\mathcal{T}_u}p_{u,j}\log p_{u,j},
    \qquad
    \overline{H}_c=\frac{1}{N_c}\sum_{u:y_u=c}H_u,
\end{equation}
where $\mathcal{T}_u$ denotes the selected experts and $p_{u,j}$ is obtained by normalizing their non-negative gate weights. The logarithm is natural; thus, under Top-3 routing, the maximum entropy is $\ln 3\approx1.099$.

Finally, the KL-based adapter activation rate (AR) measures the proportion of windows for which the adapter is activated:
\begin{equation}
    \mathrm{AR} =
    \frac{1}{N_c}\sum_{u:y_u=c}\mathbb{I}(D_u>\tau),
\end{equation}
where $D_u$ is the KL divergence between the current and reference interest distributions, $\tau$ is the activation threshold, and $N_c$ is the number of windows with ground-truth label $c$. The adapter is activated when $D_u>\tau$; otherwise, the backbone representation is used directly.

\paragraph{Expert Usage and Label--Expert Correspondence.}
Table~\ref{tab:expert_analysis} summarizes expert usage and activation statistics across the 15 interest categories. Politics and Entertainment exhibit particularly strong category--expert correspondence, with F1 scores of 91.10\% and 83.63\%, respectively. Technology and Gaming also achieve relatively high recall (76.40\% and 73.10\%). In contrast, Sports and Fashion \& Beauty are most frequently assigned to the Food and Science experts, respectively. Overall, the label--expert agreement is 72.0\%, suggesting that the routing mechanism captures category-related patterns while allowing cross-category associations rather than enforcing a strict one-to-one mapping.

\begin{table*}[h]
\centering
\small
\setlength{\tabcolsep}{4pt}
\caption{Expert routing and adapter activation statistics across interest categories.}
\label{tab:expert_analysis}
\begin{tabular*}{0.95\textwidth}{@{\extracolsep{\fill}}lccccccc}
\toprule
\textbf{Category} & \textbf{Count} & \textbf{Top Expert} &
\textbf{MG} & \textbf{Recall} & \textbf{Precision} & \textbf{H} & \textbf{AR (\%)} \\
\midrule
Entertainment & 4646 & Entertainment & 0.197 & 0.822 & 0.851 & 0.392 & 86.00\% \\
Politics & 4587 & Politics & 0.194 & 0.888 & 0.935 & 0.269 & 96.40\% \\
Daily Life & 2279 & Daily Life & 0.088 & 0.623 & 0.657 & 0.497 & 59.30\% \\
Gaming & 2157 & Gaming & 0.076 & 0.731 & 0.762 & 0.567 & 33.20\% \\
Sports & 1797 & Food & 0.000 & 0.521 & 0.478 & 0.948 & 90.20\% \\
Technology & 1783 & Technology & 0.101 & 0.764 & 0.798 & 0.471 & 35.10\% \\
Business & 1469 & Business & 0.079 & 0.512 & 0.539 & 0.667 & 23.80\% \\
Art & 985 & Art & 0.073 & 0.548 & 0.582 & 0.573 & 36.20\% \\
Health \& Fitness & 939 & Health \& Fitness & 0.017 & 0.447 & 0.473 & 0.996 & 28.40\% \\
News & 924 & News & 0.046 & 0.498 & 0.521 & 0.915 & 30.50\% \\
Fashion \& Beauty & 805 & Science & 0.000 & 0.443 & 0.412 & 1.049 & 25.20\% \\
Education & 787 & Education & 0.025 & 0.413 & 0.438 & 0.967 & 24.00\% \\
Culture & 525 & Culture & 0.013 & 0.382 & 0.405 & 1.009 & 31.80\% \\
Food & 447 & Food & 0.058 & 0.402 & 0.420 & 1.053 & 14.50\% \\
Science & 410 & Science & 0.032 & 0.365 & 0.381 & 1.066 & 19.50\% \\
\bottomrule
\end{tabular*}
\end{table*}

\paragraph{Routing Entropy and Adapter Activation.}
The overall mean routing entropy is 0.584 across 24,540 test windows, below the maximum of $\ln 3\approx1.099$. Politics (0.269) and Entertainment (0.392) exhibit relatively concentrated routing distributions, whereas Science (1.066) and Fashion \& Beauty (1.049) show more evenly distributed weights among the selected experts.

The overall adapter activation rate is 59.8\%, with substantial variation across categories. For example, Politics and Sports have relatively high activation rates of 96.40\% and 90.20\%, respectively, while Food and Science exhibit lower rates of 14.50\% and 19.50\%. These differences indicate varying levels of estimated temporal drift across interest categories.

\paragraph{Qualitative Case Studies.}

\begin{table}[h]
\centering
\small
\setlength{\tabcolsep}{4pt}
\caption{Representative cases illustrating expert routing and adapter activation.}
\label{tab:expert_cases}
\begin{tabular}{lccccc}
\toprule
\textbf{Pattern} & \textbf{User Window} & \textbf{Label} &
\textbf{Top Expert} & \textbf{Entropy} & \textbf{KL} \\
\midrule
Stable Routing & U\_001\_w011 & Entertainment & Entertainment & 0.023 & 0.104 \\
Stable Routing & U\_002\_w010 & Entertainment & Entertainment & 0.026 & 0.017 \\
High Uncertainty & U\_003\_w010 & Health \& Fitness & Health \& Fitness & 1.099 & 0.005 \\
High Uncertainty & U\_004\_w010 & Health \& Fitness & Art & 1.099 & 0.003 \\
Drift Triggered & U\_005\_w010 & Politics & Politics & 0.213 & 8.04 \\
Drift Triggered & U\_006\_w010 & Health \& Fitness & Health \& Fitness & 0.948 & 5.77 \\
\bottomrule
\end{tabular}
\end{table}

Table~\ref{tab:expert_cases} presents representative windows illustrating three routing and activation patterns. The stable-routing cases exhibit low routing entropy, with the corresponding category expert receiving the highest weight. Although their KL divergence is relatively small, both values exceed the activation threshold, and the adapter is activated. The high-uncertainty cases exhibit maximum routing entropy ($1.099$), indicating nearly uniform weights across the three selected experts. However, their KL divergence remains below the threshold, so the adapter is not activated. In contrast, the drift-triggered cases exhibit substantially higher KL divergence, leading to adapter activation. These examples highlight the complementary roles of routing entropy and KL-based activation: the former characterizes the distribution of expert weights, whereas the latter determines whether adapter fusion is applied.

\newpage

\section*{NeurIPS Paper Checklist}

\begin{enumerate}

\item {\bf Claims}
    \item[] Question: Do the main claims made in the abstract and introduction accurately reflect the paper's contributions and scope?
    \item[] Answer: \answerYes{}, 
    \item[] Justification: We accurately describe our contributions in the abstract and introduction.

\item {\bf Limitations}
    \item[] Question: Does the paper discuss the limitations of the work performed by the authors?
    \item[] Answer: \answerYes{} 
    \item[] Justification: We discuss limitations in Section ~\ref{app:disscussion}, including annotation noise, fixed window lengths, static expert count, and offline training.

\item {\bf Theory assumptions and proofs}
    \item[] Question: For each theoretical result, does the paper provide the full set of assumptions and a complete (and correct) proof?
    \item[] Answer: \answerNA{} 
    \item[] Justification: This paper does not contain theoretical result.

\item {\bf Experimental result reproducibility}
    \item[] Question: Does the paper fully disclose all the information needed to reproduce the main experimental results of the paper to the extent that it affects the main claims and/or conclusions of the paper (regardless of whether the code and data are provided or not)?
    \item[] Answer: \answerYes{} 
    \item[] Justification: The paper provides detailed descriptions of the model architecture, training procedure, dataset construction, and evaluation settings (Sections ~\ref{app:method} and ~\ref{app:experiments} ). We will release anonymized data and code to support reproducibility.

\item {\bf Open access to data and code}
    \item[] Question: Does the paper provide open access to the data and code, with sufficient instructions to faithfully reproduce the main experimental results, as described in supplemental material?
    \item[] Answer: \answerYes{} 
    \item[] Justification: We will release anonymized data and full code upon acceptance, including preprocessing scripts, model implementation, and training/evaluation pipelines. Detailed instructions for data preparation, environment setup, and running experiments will be provided in the supplemental material to ensure faithful reproduction of the main results.

\item {\bf Experimental setting/details}
    \item[] Question: Does the paper specify all the training and test details (e.g., data splits, hyperparameters, how they were chosen, type of optimizer) necessary to understand the results?
    \item[] Answer: \answerYes{} 
    \item[] Justification: We specify data splits, optimizer, batch size, and hyperparameter selection in Section ~\ref{app:experiments} and Appendix ~\ref{app:impl}.

\item {\bf Experiment statistical significance}
    \item[] Question: Does the paper report error bars suitably and correctly defined or other appropriate information about the statistical significance of the experiments?
    \item[] Answer: \answerYes{} 
    \item[] Justification: We report the mean and standard deviation over three random seeds in Appendix ~\ref{app:statistical_results}, together with statistical significance tests comparing DUMoE against the baselines.

\item {\bf Experiments compute resources}
    \item[] Question: For each experiment, does the paper provide sufficient information on the computer resources (type of compute workers, memory, time of execution) needed to reproduce the experiments?
    \item[] Answer: \answerYes{} 
    \item[] Justification: We report GPU experiments compute resources in Appendix ~\ref{app:impl}.
    
\item {\bf Code of ethics}
    \item[] Question: Does the research conducted in the paper conform, in every respect, with the NeurIPS Code of Ethics \url{https://neurips.cc/public/EthicsGuidelines}?
    \item[] Answer: \answerYes{} 
    \item[] Justification: We have reviewed and complied with the NeurIPS Code of Ethics. Data is collected from X's public API in accordance with its terms of service.

\item {\bf Broader impacts}
    \item[] Question: Does the paper discuss both potential positive societal impacts and negative societal impacts of the work performed?
    \item[] Answer: \answerYes{} 
    \item[] Justification: We discuss potential benefits and risks in Appendix ~\ref{app:impact}, along with mitigation strategies for data release.
    
\item {\bf Safeguards}
    \item[] Question: Does the paper describe safeguards that have been put in place for responsible release of data or models that have a high risk for misuse (e.g., pre-trained language models, image generators, or scraped datasets)?
    \item[] Answer: \answerYes{} 
    \item[] Justification: We will release only anonymized user IDs and tweet IDs, and will apply content filtering to remove harmful posts prior to public release.

\item {\bf Licenses for existing assets}
    \item[] Question: Are the creators or original owners of assets (e.g., code, data, models), used in the paper, properly credited and are the license and terms of use explicitly mentioned and properly respected?
    \item[] Answer: \answerYes{} 
    \item[] Justification: We cite all baseline models and the encoder with their original references. X data is collected under the platform's academic access terms.

\item {\bf New assets}
    \item[] Question: Are new assets introduced in the paper well documented and is the documentation provided alongside the assets?
    \item[] Answer: \answerYes{} 
    \item[] Justification: We document the dataset construction, domain taxonomy, and annotation protocol in Section ~\ref{app:dataset construction} and Appendix ~\ref{app:dataset}. 

\item {\bf Crowdsourcing and research with human subjects}
    \item[] Question: For crowdsourcing experiments and research with human subjects, does the paper include the full text of instructions given to participants and screenshots, if applicable, as well as details about compensation (if any)? 
    \item[] Answer: \answerNA{} 
    \item[] Justification: The dataset is constructed from publicly available social media data and primarily annotated automatically using GPT-5.1. Manual inspection by research team members is conducted on a small subset for annotation quality verification. No crowdsourcing experiments or human subject studies are conducted.

\item {\bf Institutional review board (IRB) approvals or equivalent for research with human subjects}
    \item[] Question: Does the paper describe potential risks incurred by study participants, whether such risks were disclosed to the subjects, and whether Institutional Review Board (IRB) approvals (or an equivalent approval/review based on the requirements of your country or institution) were obtained?
    \item[] Answer: \answerNA{} 
    \item[] Justification: We only use publicly available social media data for human annotation quality checks, which does not require IRB approval under our institutional guidelines.

\item {\bf Declaration of LLM usage}
    \item[] Question: Does the paper describe the usage of LLMs if it is an important, original, or non-standard component of the core methods in this research? Note that if the LLM is used only for writing, editing, or formatting purposes and does \emph{not} impact the core methodology, scientific rigor, or originality of the research, declaration is not required.
    \item[] Answer: \answerYes{} 
    \item[] Justification: GPT-5.1 is used in the pseudo-label annotation pipeline (Section ~\ref{app:dataset construction}), directly affecting dataset construction and experimental results.

\end{enumerate}


\end{document}